\pdfoutput=1

\documentclass[11pt]{article}
\usepackage{authblk}
\usepackage{acl}

\usepackage{times}
\usepackage{latexsym}

\usepackage[T1]{fontenc}

\usepackage[utf8]{inputenc}

\usepackage{microtype}
\usepackage{graphicx}
\usepackage{xcolor}
\usepackage{todonotes}
\usepackage{multirow}
\colorlet{shadecolor}{blue!20}

\newcommand{\modelname}{\textsc{Ecc}}
\title{Improving the Faithfulness of Abstractive Summarization via Entity Coverage Control}

\author[1]{Haopeng Zhang}
\author[2]{Semih Yavuz}
\author[2]{Wojciech Kryscinski}
\author[2]{Kazuma Hashimoto}
\author[2]{Yingbo Zhou}
\affil[1]{IFM Lab, University of California,Davis}
\affil[2]{Salesforce Research}
\affil[1]{haopeng@ifmlab.org}
\affil[2]{\{syavuz,wojciech.kryscinski,k.hashimoto,yingbo.zhou\}@salesforce.com}

\begin{document}
    \maketitle

\begin{abstract}
Abstractive summarization systems leveraging pre-training language models have achieved superior results on benchmark datasets.
However, such models have been shown to be more prone to hallucinate facts that are unfaithful to the input context. 
In this paper, we propose a method to remedy entity-level extrinsic hallucinations with \textbf{E}ntity \textbf{C}overage \textbf{C}ontrol (\modelname). 
We first compute entity coverage precision and prepend the corresponding control code for each training example, which implicitly guides the model to recognize faithfulness contents in the training phase.
We further extend our method via intermediate fine-tuning on large but noisy data extracted from Wikipedia to unlock zero-shot summarization.
We show that the proposed method leads to more faithful and salient abstractive summarization in supervised fine-tuning and zero-shot settings according to our experimental results on three benchmark datasets XSum, Pubmed, and SAMSum of very different domains and styles. 

\end{abstract}
    \section{Introduction}
Abstractive summarization aims to generate a compact and fluent summary that preserves the most salient content of the source document. Recent advances in pre-trained language models~\cite{devlin2018bert,liu2019text,lewis-etal-2020-bart} have led to improvements in the quality of generated summaries.

However, one prominent limitation of existing abstractive summarization systems is the lack of faithfulness of generated outputs. Faithful summaries should only contain content that can be derived from the source document instead of hallucinated or fabricated statements. Summary hallucination could be categorized by the information source as intrinsic and
extrinsic hallucinations. \citet{cao2018faithful, kryscinski2019neural} showed that about 
 $30\%$ of the summaries generated by seq2seq models suffer from the hallucination phenomenon at either the entity level or the summary level. Table ~\ref{intro_ex} shows an example of a model generated
summary with hallucinated entities. The BBC article discusses a teenage science competition streamed on the Youtube website, while a BART-based summarizer makes up the term 'Gumtree' instead. Such hallucinations may cause factual errors and hinder the practical use of summarization models.

\begin{table}[t]
\small
\begin{tabular}{l}
\begin{tabular}[c]{@{}l@{}} \textbf{Source:} \textit{When the experiments are eventually run, the }\\\textit{results will be streamed live on {\color{orange}YouTube}. Alongside Prof} \\ \textit{{\color{cyan} Hawking}, the judging panel consists of [...]}\end{tabular}\\
\hline
\begin{tabular}[c]{@{}l@{}} \textbf{Summary:} \textit{{\color{cyan} Stephen Hawking} joined the judging panel of a}\\\textit{science competition on the internet education site} \textit{\textbf{\color{red}Gumtree}}.\end{tabular}\\
\hline
\end{tabular}

\caption{An example of model generated unfaithful summary due to entity hallucination from XSum dataset.}
\label{intro_ex}
\end{table}
 Faithfulness and factuality in abstractive summarization has received growing attention from the NLP community \cite{ kryscinski-etal-2020-evaluating,goyal-durrett-2021-annotating,zhu-etal-2021-enhancing,narayan2021planning}.
Recent works have attempted to address the hallucination problem at the entity level by reducing hallucinated entities during generation. \citet{chen2021improving} proposed a post-processing method, which replaces the hallucinated entities in the generated outputs with the same type entities in the source document. However, it introduces additional errors to the summary and increases the intrinsic hallucination. \citet{nan-etal-2021-entity} proposed to address entity hallucination by filtering the training data and multi-task learning with summary-worthy named-entities classification. However, the method sacrifices part of the training data and decreases the quality of the summary.

To address the above issues, we propose to solve entity hallucination by guiding the model learning process with entity control code (\modelname) \cite{keskar2019ctrl,he2020ctrlsum, fan2017controllable}. We utilize the entity coverage precision between the training document and its reference summary as faithfulness guidance and prepend it to the corresponding document in the training phase. Then, we prepend faithful control code during inference and reduce hallucinated entities effectively without decreasing the fluency and salience of generated summaries according to our experimental results. In addition, we extend control code to a Wikipedia-based intermediate fine-tuning model, which generates faithful and salient summaries across domains in the zero-shot setting. We validate our methods on three benchmark datasets across different domains, and experimental results demonstrate the effectiveness of our methods.
    \section{Methods}
\begin{figure}
    \centering
    \includegraphics[width = 0.5\textwidth]{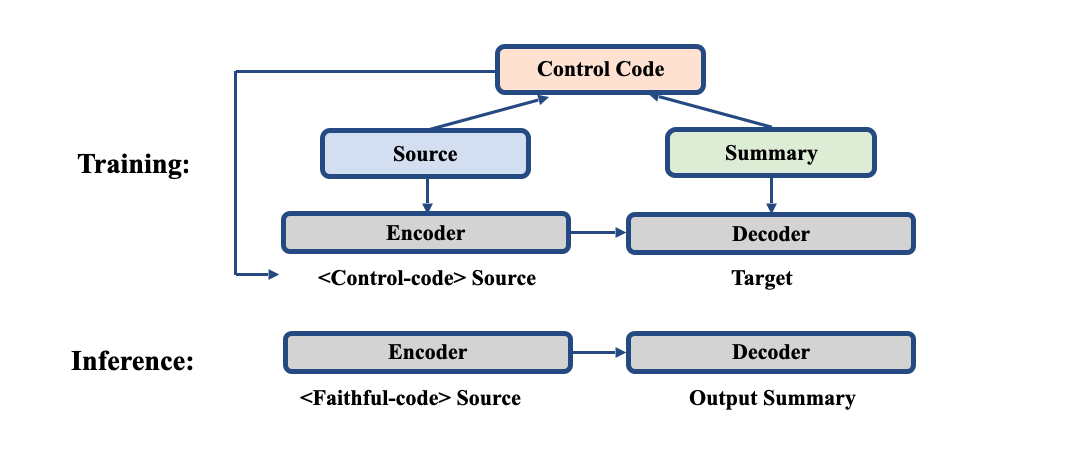}
    \caption{Entity Coverage Control for seq2seq model.}
    \label{code_figure}
\end{figure}


\subsection{Problem Formulation}
Let $D =\{(d_1,s_1) , (d_2,s_2), . . . , (d_n,s_n)\}$ denote a dataset composed of $n$ document and summary pairs. During inference phase, a seq2seq model generates summary hypothesis $h_i$ for a given document $d_i$ by  computing the probability $p_{\theta}(h_i|d_i)$. The generated summary $h_i$ is expected to be faithful, which means all the information in $h_i$ should be entailed by the source document $d_i$.

Following \cite{nan-etal-2021-entity}, we quantify entity-level hallucination with entity coverage precision $\mathbf{prec_{en}}$. 
It approximates the faithfulness by measuring the ratio of the named entities in the summary that are coming from the source document. Formally, it is defined as:
\begin{equation}
    \mathbf{prec_{en}} = \left| \mathcal{N}(h) \cap \mathcal{N}(s) \right|/ \left|\mathcal{N}(h) \right|
\end{equation}
where $\mathcal{N}(t)$ represents the set of all named entities found in a given input text $t$. 

\subsection{Entity Coverage Control}
Figure ~\ref{code_figure} shows our entity coverage control method. 
We generate a control code $C_i$ for each training document and reference summary pair $(d_i,s_i)$ so the seq2seq model generates a summary conditioned on both the source document $d_i$ and its control code ${C_i}$, which is represented as $p_{\theta}(h_i|d_i,C_i)$.

We first compute entity coverage precision $\mathbf{prec_{en}}$ for each document and reference summary pair $(d_i,s_i)$ in the training set $D$. Then, we quantize $\mathbf{prec_{en}}$ into $k$ discrete bins, each representing a range of entity faithfulness. These bin boundaries are selected to ensure that each bin contains roughly the same number of training examples to avoid data imbalance. We then represent each bin by a special token control code ${C_i}$ and add all these special tokens $\{C_1, C_2, . . . , C_k\}$ to the input vocabulary of our seq2seq model.

During training, we prepend the corresponding pseudo label ${C_i}$ to the input document as control code. The seq2seq model is now conditioned on both the source document $d_i$ and its control code ${C_i}$, so it could learn different faithful level generation patterns from the control codes. Then during inference, we prepend the high faithfulness control code ${C_k}$ to all documents in the test set and generate faithful summaries by $p_{\theta}(h_i|d_i,C_k)$.

\subsection{Controllable Intermediate Fine-tuning}

Large pre-trained language models \cite{devlin2018bert,lewis2019bart} perform poorly in the zero-shot summarization setting since sentence salience information is not learned through pre-training tasks \cite{zhang2020pegasus}. Thus, we propose a controllable generalized intermediate fine-tuning for zero-shot summarization. 

We first generate pseudo document summary pairs from Wikipedia article dump with similar summary length ($n$), document length ($m$) and abstractiveness ($a$) to the target datasets following Wikitransfer \cite{fabbri-etal-2021-improving}. Instead of training different models for different target datasets as in WikiTransfer, we propose a unified model that generalizes well across different domains. Assume we have $l$ target-specific pseudo training subsets $\{D_1(n_1,m_1,a_1), . . . , D_l(n_l,m_l,a_l)\}$, we give each subset another special token ${E_i}$ as a pseudo label to represent the target-specific pattern and also add all these special tokens $\{E_1, E_2, . . . , E_l\}$ to the input vocabulary of the seq2seq model. In the training phase, we prepend the corresponding target code ${E_i}$ to the document, and a summary is generated conditioned on both the source document $d_i$ and its target control code ${E_i}$, which is represented as $p_{\theta}(h_i|d_i,E_i)$. This allows for control over the domain and generation style of generated summaries by prepending different domain control codes during inference. The control codes are also stackable, so we can stack the target control with entity coverage control for faithful zero-shot summarization, which could be denoted as $p_{\theta}(h_i|d_i,C_i,E_i)$. 
\begin{table}
\centering
\small
\begin{tabular}{l|c|ccc} 

\multicolumn{5}{c}{\textbf{Pubmed}}\\
\hline { Model } &\begin{tabular}{@{}c@{}}Entity \\ Precision\end{tabular} & R-1 & R-2 & R-L\\
\hline
{Reference} &42.85  &100  &100&100  \\
{BART}$_{large}$ &74.31  &43.35 &16.20   & 39.50  \\
\modelname  & {\textbf{76.38}}  &\textbf{43.46}  &\textbf{16.24}   &\textbf{39.68} \\
\hline
\multicolumn{5}{c}{\textbf{SAMSum}}\\
\hline { Model } &\begin{tabular}{@{}c@{}}Entity \\ Precision\end{tabular} & R-1 & R-2 & R-L\\
\hline
{Reference} &71.20 &100  &100&100 \\
{BART}$_{large}$ &78.50  &52.39 &\textbf{27.89}   &\textbf{43.58} \\
\modelname &\textbf{80.23}   &\textbf{52.42}  & 27.69  &43.34 \\
\hline

\end{tabular}
\caption{Experiment results in the supervised fine-tuning setting on Pubmed and SAMsum datasets, XSum results are reported in Table ~\ref{base} }
\label{super}
\end{table}
\begin{table}
\centering
\small

\begin{tabular}{l|cc|cc}  \multicolumn{5}{c}{\textbf{XSum}}\\
\hline 
{ Model } &\begin{tabular}{@{}c@{}}Entity \\ Precision\end{tabular} & FEQA & R-1 & R-L \\
\hline
\textsc{Bart} &54.11 &22.50  &\textbf{44.78}    &\textbf{36.64}\\
\textbf{+Correction}  &55.57 &25.62 &43.48    &35.32\\
\textbf{+Filter} &\textbf{70.49} &\textbf{26.73} &42.19    &33.97\\
\hline
\modelname  &59.38 &26.51  &43.82     &35.97\\
\hline

\end{tabular}

\caption{Performance comparison against state-of-the-art baselines on XSum dataset.}
\label{base}
\end{table}

    \section{Experiments}
\subsection{Experiment Settings}
\paragraph{Datasets, evaluation and implementation:} 

We experiment with three summarization datasets in different domains: news dataset {\textit{XSum}} \cite{narayan2018don}, scientific paper dataset {\textit{Pubmed}} \cite{cohan2018discourse}, and dialogue dataset {\textit{Samsum}} \cite{gliwa2019samsum}. We use \textit{Entity Precision} \cite{nan-etal-2021-entity} and \textit{FEQA} \cite{durmus-etal-2020-feqa} to evaluate summary faithfulness and use \textit{ROUGE} \cite{lin2004rouge} to evaluate the fluency and salience. 
We also ask expert annotators to perform a human evaluation in both summary faithfulness and quality. We use BART-large as backbone model and set hyperparameter $k=3$ for all experiments. The three discrete {\modelname} bins are represented with control codes: <FF-low>, <FF-mid> and <FF-high> respectively. More implementation details are described in Appendix ~\ref{sec:imple}.

\paragraph{Baselines:} We compare our methods with two state-of-the-art methods in summarization faithfulness: (1)Post-processing \textbf{correction} in \cite{chen2021improving} (2)Entity-based data \textbf{filtering} in \cite{nan-etal-2021-entity} together with original \textbf{BART}. For zero-shot summarization, we compare with state-of-the-art method \textbf{WikiTransfer} \cite{fabbri-etal-2021-improving}.

\begin{table}
\centering
\small
\begin{tabular}{l|c|ccc}  \multicolumn{5}{c}{\textbf{Xsum}}\\
\hline 
{ Model } &\begin{tabular}{@{}c@{}}Entity \\ Precision\end{tabular}  & R-1 & R-2 & R-L\\
\hline
\textsc{Bart} &\textbf{92.61}   &19.45 & 3.01   &13.29 \\
\textsc{Wikitransfer}  &50.50  &29.39 &8.90   &21.98 \\
\hline
\modelname-zero  &55.48   &\textbf{30.05}  &\textbf{9.72}   &\textbf{22.99} \\
\hline
\multicolumn{5}{c}{\textbf{Pubmed}}\\
\hline 
{ Model } &\begin{tabular}{@{}c@{}}Entity \\ Precision\end{tabular}  & R-1 & R-2 & R-L\\
\hline
\textsc{Bart} &42.85   &31.65&10.17   &16.60 \\
\textsc{Wikitransfer} &62.72  &\textbf{38.64} &13.28   &\textbf{19.37} \\
\hline
\modelname-zero  &\textbf{68.13}   &38.42  &\textbf{13.34}   &19.32 \\
\hline

\end{tabular}
\caption{Model performance in the zero-shot summarization setting.}
\label{zero}
\end{table}

\begin{table}
\small
\centering
\
\begin{tabular}{l|cccc} 

{ Model } &Faith. \% & Ex. \% & In. \% &Quality\\
\hline
\textsc{Bart} &  15.0 & 54.0& 39.0  &2.31 \\
\textbf{+Correction} &27.0  &48.0 &57.0  &2.42 \\
\modelname  &28.0  &\textbf{41.0}    & \textbf{37.0} &\textbf{2.43} \\
\modelname-zero  &\textbf{31.0}   &48.0 & 38.0  &1.73 \\
\hline

\end{tabular}
\caption{Human evaluation results of 50 test examples sampled from XSum dataset. Results with inter-annotator agreement are reported in Appendix ~\ref{sec:confidence}. }
\label{human}
\end{table}

\subsection{Automatic Evaluation}
Table ~\ref{super} shows the performance of {\modelname} in the supervised setting. Compared to the summaries generated by BART, our method increases the entity coverage precision significantly with roughly the same summary quality. Table ~\ref{base} shows the performance comparison to strong baselines on the XSum dataset. Our methods achieves comparable faithfulness improvements without degrading the summary quality compared to data filtering and post-processing methods. We notice there is a trade-off between entity coverage precision and summary quality in Xsum dataset, which is likely due to the low faithfulness level of the reference summaries of Xsum \cite{maynez-etal-2020-faithfulness}.

Table ~\ref{zero} shows the zero-shot summarization results. We notice BART tends to copy from the source document, so it achieves high entity coverage precision ($92.61$) but low summary quality. In contrast, with our intermediate fine-tuning, BART learns the characteristic of the downstream dataset and achieves a considerable improvement in ROUGE score. Compared to the baseline Wikitransfer, we see improvements in both the entity coverage precision and summary quality. Our model is also generalized cross datasets, so we use one model for different downstream targets instead of training separate models like Wikitransfer.

\subsection{Human Evaluation}
Table ~\ref{human} shows the human evaluation results on
the 50 randomly sampled subset of articles from the XSum dataset following the setting of \cite{chen2021improving}. Four expert annotators assign each summary output into three faithfulness categories (faithful summary, intrinsic hallucination, extrinsic hallucination) and three summary quality categories (low(1), medium (2), high(3)). Note that a summary may contain both intrinsic and extrinsic hallucinations. As the results show, our {\modelname} model improves the faithfulness of the summaries without degrading summary quality, which agrees with our automatic evaluation results.

\begin{figure}
    \centering
    \includegraphics[width=0.5\textwidth]{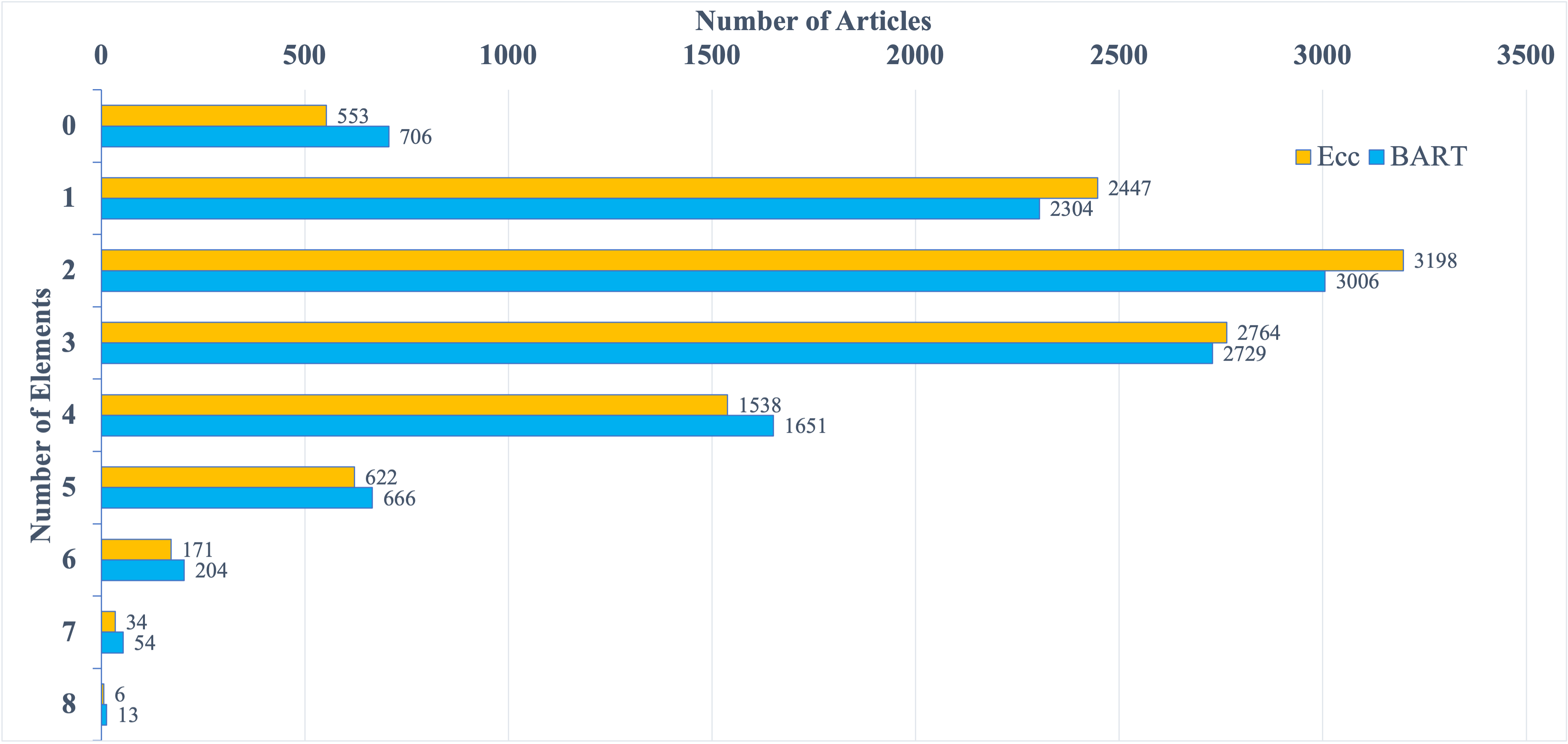}
    \caption{Number of entities in the generated summary from BART and \modelname .}
    \label{entity_figure}
\end{figure}
\begin{table}
\centering
\small
\begin{tabular}{l|c|ccc} 

\hline 
{ Model } &\begin{tabular}{@{}c@{}}Entity \\ Precision\end{tabular}  & R-1 & R-2 & R-L\\
\hline
{BART}$_{large}$ &54.11  &\textbf{44.78} &21.60
   &36.64 \\
\textsc{Low}  &51.32   &44.03  &21.23   &36.12 \\
\textsc{Medium}  &53.50  &43.94 &21.21  &35.94 \\
\textsc{High}  &{\textbf{59.38}}  &43.82 &21.15  &35.97 \\
\hline

\end{tabular}
\caption{Comparison of summaries docoding with different control codes on XSum Dataset.}
\label{inference}
\end{table}

\begin{table}
\small
\centering
\begin{tabular}{l}
\hline
\textbf{Document:} Saints captain \textbf{<mask>} Anderson claims he \\was punched by Kiernan during last week's 1-1 draw \\ between  the sides. [...] 
\\
\textbf{Bart:} St Johnstone's \textit{\textbf{\color{red}Gary}} Anderson says Rangers mid-\\fielder \textit{\textbf{\color{red}John }}Kiernan should face a Scottish FA disciplinary \\hearing over an alleged punch. \\
\hline
\textbf{Reconstructed <mask>} from 1st sentence context:\\ 
Top-5: ['Paul', 'Mark', 'Tom', 'James', 'Ryan’] \\
\textbf{Reconstructed <mask>} from full source context:\\ 
Top-5: ['Craig', '\textit{\textbf{\color{red}Gary }}', 'Kier', '\textit{\textbf{\color{blue}Steven}}', 'Anderson’] \\

\hline

\end{tabular}
\caption{An example of hallucinated entity analysis with mask token refilling by BART. The ground truth is 'Steven Anderson' according to web search.}
\label{error}
\end{table}
    
\section{Analysis and Discussion}

\paragraph{\textbf{Does our model generate fewer entities to be safe?}} One obvious way to get higher entity coverage precision is to avoid generating entities or generating extra non-sense named entities from the source document. We show the distribution of the number of entities in the generated summaries by our model and BART in Fig ~\ref{entity_figure}. We see that the two distributions are very similar and have almost the same mean number of entities. As a result, we argue that our method doesn't under-generate nor over-generate entities from the source document, and we don't need to separately control the entity compression rate.

\paragraph{\textbf{How does control code affect inference phase?}}  We also study the effect of decoding with different control codes. We prepend different entity coverage control codes during inference on the XSum test set. As shown in Table ~\ref{inference}, our model still generates reasonable summaries when inferred with low and medium control codes. We notice that summaries inferred with low control codes have higher ROUGE scores, which agrees with the trade-off described earlier.\\



\noindent{\textbf{Why does BART generate hallucinated tokens?}} \\
As shown in Table ~\ref{error}, fine-tuned BART generates `Gary Anderson' according to the context `Saints captain Anderson' , which is erroneous since the actual captain is `Steven Anderson'. Language models contain abundant relational knowledge from pre-training data and could be extracted by masked text filling \cite{petroni2019language}. Similarly, we insert a mask token before `Anderson' and probe untuned BART to fill the masked tokens. BART generates `Paul Anderson' (actor) when only given the first sentence context. When given the whole news article, BART learns the context is sports-related and generates famous athletes `Craig Anderson' (hockey athlete) and `Gary Anderson' (football athlete) according to its pre-trained prior knowledge. The ground truth `Steven Anderson' appears much less frequent during pre-training, so BART has a low probability of generating it correctly. We observe the same for ground truth `Rob Kiernan', which probably appears less frequently in BART's pre-training corpus.



    \section{Related Work}
 The faithfulness and factuality in abstractive summarization has received growing attention by the summarization community recently \cite{ kryscinski-etal-2020-evaluating,cao2018faithful,goyal-durrett-2021-annotating}. \citet{maynez-etal-2020-faithfulness} categorized hallucinations by the information source as intrinsic and extrinsic hallucinations. Researchers have turned to textual entailment \cite{maynez-etal-2020-faithfulness}, question answering (QA)\cite{durmus-etal-2020-feqa,wang-etal-2020-asking}, Natural Language Inference (NLI) \cite{kryscinski-etal-2020-evaluating} and entity level precision \cite{nan-etal-2021-entity} for automatic faithfulness evaluation. To improve the faithfulness of generated summaries. \citet{cao2018faithful} proposes a fact-aware summarization model with open information extraction and dependency parse technologies. \cite{zhu-etal-2021-enhancing} uses graph attention to integrate factual
relations into the summary generation process. Recent works also focus on addressing entity-level hallucination problems. \citet{chen2021improving} proposes a post-processing method to correct hallucinated entities and \cite{nan-etal-2021-entity} addresses entity hallucination by filtering the training data and multi-task learning. One concurrent work \citet{narayan2021planning} incorporates entity chain content planning to guide faithful summary generation.

The transformer-based seq2seq architecture \cite{vaswani2017attention} currently
dominates the state-of-the-art performance in many
NLP tasks \cite{liu2019text,zhang2020graphbert,zhang2020text}. We use BART \cite{lewis-etal-2020-bart} as a backbone for abstractive summarization in this work, but our method is generally appliable for all seq2seq models.

Our work is also related to controllable abstractive summarization. \citet{liu2018controlling} controls the summary length by extending
a convolutional sequence to sequence model. \citet{he2020ctrlsum} introduces a keyword guided framework for entity-centric, length-controllable summarization and question-guided
summarization. \citet{fan2017controllable} proposes to control the summary generation with a list of desired named entities. Recently, \citet{feng2021language} proposes to use language models to generate pseudo labels to control the generation of dialogue summarization. Our work uses control code to improve summary generation faithfulness and cross-domain generalizability.
    \section{Conclusion}
In this paper, we propose {\modelname} to address extrinsic hallucination in abstractive summarization in both supervised and zero-shot settings. 
Our extensive experiments demonstrate that the proposed method effectively reduces entity hallucination without hurting the quality of the generated summaries.
\section*{Acknowledgement}
Work done during internship at Salesforce Research.  Thanks to Man Luo, Xi Ye, and everyone at Salesforce Research for helpful discussions, as well as to the
anonymous reviewers for their helpful feedback.
    
    \bibliography{anthology,custom}
    \bibliographystyle{acl_natbib}
    
    \newpage
    \appendix
    \label{sec:appendix}

\section{Implementation Details}
\label{sec:imple}

We use Huggingface libraries \cite{wolf-etal-2020-transformers} for all our experiment implementations. Our backbone abstractive summarization model is
BART-large \cite{lewis-etal-2020-bart}, a pre-trained denoising autoencoder language model with $336M$ parameters based on the sequence-to-sequence transformer \cite{vaswani2017attention}. For fair comparison, we fine-tune BART-large on each dataset for on 8 Tesla A100 GPU pods with same learning rate $5e-5$ with weight decay using Adam optimizer \cite{kingma2014adam}. We set hyperparameter $k=3$ for all experiments. Larger number of $k$ doesn't increase the performance significantly. The three discrete {\modelname} bins are represented with control codes: <FF-low>, <FF-mid> and <FF-high> respectively. The entity coverage precision boundaries are $0.36$ and $0.5$ for Pubmed, $0.33$ and $0.66$ for SAMsum and Xsum.

For entity recognition, we use a neural Named Entity Recognition (NER) system from the Stanza NLP toolkit \cite{qi-etal-2020-stanza}
trained on the OntoNotes corpus \cite{weischedel2011ontonotes} except for Pubmed dataset. Since Pubmed is a medical scientific article collection, we use biomedical, scientific, and clinical text Named Entity Recognition toolkit scispaCy \cite{neumann-etal-2019-scispacy} instead.

\begin{table*}
\small
\begin{tabular}{l}
\hline
 \textbf{BART:} A video game based on one of the world's most popular wrestling traditions has been launched at the E3 \\gaming show in Los Angeles.'\\
 \textbf{Correction:} A video game based on one of the world's most popular wrestling traditions has been launched at the \\E3 gaming show in Mexico.\\
 \textbf{ECC:} A video game dedicated to Mexican wrestling has been released at E3.\\
 \textbf{Reference:} One of the more unusual titles at E3, the worlds largest video games exhibition held each year in \\Los Angeles, is Konami's Lucha Libre AAA: Heroes del Ring.\\
\hline
\textbf{BART:} Tourists in Spain have been accused of harassing a dolphin after it became stranded on a beach.\\
\textbf{Low Code:} A dolphin that became stranded in the sea off the coast of Spain has been harassed by a group \\of tourists.\\
\textbf{High Code:} A dolphin that became stranded in the sea off the coast of Andalucia has been harassed by tourists.\\
\textbf{Reference:} A baby dolphin has died after it was surrounded by tourists looking to take photographs on a beach \\in southern Spain.\\
\hline
\textbf{Document:} The warning begins at 22:00 GMT on Saturday and ends at 10:00 on Sunday.
The ice could lead to \\difficult driving conditions on untreated roads and slippery conditions on pavements, the weather service warned.\\
Only the southernmost counties and parts of the most westerly counties are expected to escape.
Counties expected \\to be affected are Carmarthenshire, Powys, Ceredigion, Pembrokeshire, Denbighshire, Gwynedd, Wrexham, \\Conwy, Flintshire, Anglesey, ..., Rhondda Cynon Taff and Torfaen.\\
\textbf{Reference:}The Met Office has issued a yellow weather warning for ice across most of Wales.\\
\hline
 \\

\end{tabular}
\caption{Representative examples from the XSum test set.}
\label{study}
\end{table*}
\section{Representative Examples Analysis }
In Table ~\ref{study}, we provide several representative examples from XSum dataset. Example 1 (first row) shows how our entity control method gets rid of hallucination terms from BART output. The reference summary here is not faithful since `Los Angeles' is not covered in the source document. The correction baseline changes `Los Angeles' to `Mexico', which is a factual error. In contrast, the \modelname output is totally faithful to the source document and contains salient information.

Example 2 (second row) shows the outputs decoded with different control codes during inference. We can see the output decoded with low faithfulness control code is still fluent and reasonable, but contains less faithful entities compared to the output decoded with high faithfulness control code.

Example 3 (third row) shows an example of factual statement, which is verifiable in the real world independent of the source text. The reference summary uses `most of Wales' to summarize the county names in the source document. This type of hallucination needs more external knowledge and commonsense reasoning to decide its factuality. Our method only focuses on entity level hallucination problems instead.

\section{Human Evaluation Confidence}
\label{sec:confidence}
\begin{table}
\small
\centering
\
\begin{tabular}{l|cccc} 

\hline 
{ Model } &Faith. \% & Ex. \% & In. \% &Quality\\
\hline
\textsc{Bart} &  $15.0\pm7.4$ & $54.0\pm 11.2$& $39.0\pm5.8 $  &$2.31\pm 0.14$ \\
\modelname  &$28.0\pm6.2$  &$41.0\pm7.2$    & $37.0\pm8.3$ &$2.43\pm0.17$ \\
ECC-zero  &$31.0\pm 2.8$   &$48.0\pm9.3$ & $38.0\pm7.2$  &$1.73\pm0.07$\\
\hline

\end{tabular}
\caption{Human evaluation results of 50 test examples sampled from XSum dataset.}
\label{t:conf}
\end{table}

Our human evaluation follows the setting of prior work \cite{chen2021improving}. We calculate the inter-annotator agreement with additional annotations from two other experts.  We estimate the adjusted mean
and $95\%$ confidence interval from the mean and
standard deviation. The full results are shown in Table ~\ref{t:conf}.
    \clearpage

\end{document}